\documentclass{article}

\usepackage{array}     
\usepackage{microtype}
\usepackage{graphicx}
\usepackage{subfigure}
\usepackage{booktabs} 
\usepackage{tabularx}

\usepackage{tikz}
\definecolor{TUred}{RGB}{165,30,55}
\definecolor{TUgold}{RGB}{180,160,105}
\definecolor{TUdark}{RGB}{50,65,75}
\definecolor{TUgray}{RGB}{175,179,183}

\definecolor{TUdarkblue}{RGB}{65,90,140}
\definecolor{TUblue}{RGB}{0,105,170}
\definecolor{TUlightblue}{RGB}{80,170,200}
\definecolor{TUlightgreen}{RGB}{130,185,160}
\definecolor{TUgreen}{RGB}{125,165,75}
\definecolor{TUdarkgreen}{RGB}{50,110,30}
\definecolor{TUocre}{RGB}{200,80,60}
\definecolor{TUviolet}{RGB}{175,110,150}
\definecolor{TUmauve}{RGB}{180,160,150}
\definecolor{TUbeige}{RGB}{215,180,105}
\definecolor{TUorange}{RGB}{210,150,0}
\definecolor{TUbrown}{RGB}{145,105,70}

\usepackage{hyperref}
\usepackage[accepted]{icml2023}

\usepackage{enumitem}
\usepackage{stmaryrd}
\usepackage{amsmath}
\usepackage{amssymb}
\usepackage{mathtools}
\usepackage{amsthm}
\usepackage[justification=centering]{caption}
\usepackage[capitalize,noabbrev]{cleveref}

\theoremstyle{plain}

\theoremstyle{definition}

\theoremstyle{remark}

\usepackage[textsize=tiny]{todonotes}

\graphicspath{ {./figures/} }
\usepackage{float}
\usepackage{verbatim} 
\usepackage{apalike}
\usepackage{amsfonts}
\usepackage{booktabs}
\usepackage{setspace}
\usepackage{amsmath}
\usepackage{csquotes}
\restylefloat{figure}
\restylefloat{table}
\usepackage[font=normalsize,skip=7pt]{caption}
\usepackage{changepage} 

\usepackage{xcolor}
\usepackage{colortbl}

\usepackage{hyperref}

\begin{document}


\twocolumn[
\icmltitle{A Comprehensive Analysis of Static Word Embeddings for Turkish}

\begin{icmlauthorlist}
\icmlauthor{Karahan Sarıtaş}{xxx}
\icmlauthor{Cahid Arda Öz}{xxx}
\icmlauthor{Tunga Güngör}{xxx}
\end{icmlauthorlist}

\icmlaffiliation{xxx}{Boğaziçi University, Computer Engineering Department, Bebek, 34342, İstanbul, Turkey}

\vskip 0.3in
]
\printAffiliationsAndNotice{} 

\begin{abstract} Word embeddings are fixed-length, dense and distributed word representations that are used in natural language processing (NLP) applications. There are basically two types of word embedding models which are non-contextual (static) models and contextual models. The former method generates a single embedding for a word regardless of its context, while the latter method produces distinct embeddings for a word based on the specific contexts in which it appears. There are plenty of works that compare contextual and non-contextual embedding models within their respective groups in different languages. However, the number of studies that compare the models in these two groups with each other is very few and there is no such study in Turkish. This process necessitates converting contextual embeddings into static embeddings. In this paper, we compare and evaluate the performance of several contextual and non-contextual models in both intrinsic and extrinsic evaluation settings for Turkish. We make a fine-grained comparison by analyzing the syntactic and semantic capabilities of the models separately. The results of the analyses provide insights about the suitability of different embedding models in different types of NLP tasks. We also build a Turkish word embedding repository comprising the embedding models used in this work, which may serve as a valuable resource for researchers and practitioners in the field of Turkish NLP. We make the word embeddings, scripts, and evaluation datasets publicly available.\end{abstract}

\section{Introduction}
\textit{Word Embeddings} for text can be defined as unsupervisedly learned word representation vectors whose relative similarities correlate with the semantic similarities of the words \cite{joulin2016bag}. By capturing the semantic relationships between words, word embeddings can be utilized in various applications including part-of-speech (PoS) tagging \cite{collobert2011natural}, question-answering \cite{xiong2017dynamic}, and named entity recognition (NER) \cite{pennington2014glove}. The history of word embeddings began with the introduction of latent semantic analysis (LSA) and latent Dirichlet allocation (LDA) in the late 1990s and early 2000s. These methods used matrix factorization and probabilistic modeling, respectively, to create word representations. In comparison to prior representation methods such as tf-idf and n-gram used in information retrieval, learning a distributed representation for words has proved to enhance performance in several natural language processing (NLP) tasks \cite{bengio2003neural, schwenk2007continuous, mikolov2011empirical}. Past these strategies, in a later work, \cite{collobert2008unified} revealed the power of word embeddings in downstream NLP tasks and introduced a neural network architecture that has influenced many of the current approaches. One of the main advantages of distributed word representations over previous statistical measures is that they provide us with the opportunity to utilize semantic and syntactic similarities between the words. 

\hspace{1mm} The recent and rapid expansion and affordability in computational power enabled researchers to train relatively more complicated architectures using neural networks. In spite of its long-time existence, learning word embeddings with neural networks gained its popularity due to Word2Vec, a word embedding toolkit that can train vector space models faster than previous approaches, which was developed by \cite{mikolov2013efficient}. Since then, numerous models have been proposed that build on the original Word2Vec architectures. \cite{bojanowski2016enriching} developed FastText, which improved upon the Word2Vec model by representing words as character $n$-grams. This modification aimed at addressing out-of-vocabulary words and improving the performance for languages with complex morphology, such as Turkish and Finnish. \cite{pennington2014glove} proposed the GloVe method which utilizes global word-word co-occurrence counts and thus effectively leverages statistics during training. In contrast to Word2Vec and FastText, GloVe uses a global co-occurrence matrix along with local context window methods. 

\hspace{1mm} Despite their relatively high performance on different NLP tasks, one of the problems with these word embedding models is that these embeddings assign a fixed vector representation to a word regardless of its context, which may not accurately capture its various meanings in different contexts. For example, the word \textit{yaz} in Turkish has different meanings and therefore should have different word embeddings in these two sentences: \textit{Numarayı bir deftere yaz} (\enquote{Write the number on a notebook} - \textit{yaz}: \enquote{(to) write}) and \textit{Bu yaz tatile cıkacagim} (\enquote{I will go on vacation this summer} - \textit{yaz}: \enquote{summer}). To overcome this issue and contextualize the word embeddings, several models have been developed such as CoVe \cite{mccann2017learned}, ELMo \cite{peters2018deep}, and BERT \cite{devlin2018bert}. ELMo contextualizes the word embeddings by training a bidirectional long short-term memory (LSTM) language model on a large corpus of text. ELMo embeddings have been shown to outperform traditional static word embeddings in a wide range of NLP tasks, such as sentiment analysis and named entity recognition \cite{peters2018deep}. BERT is a transformer-based language model that aims to make it easier to adapt to different NLP tasks by fine-tuning the pre-trained model without complicated modifications to the architecture.

Although previous studies have explored word embedding models for NLP tasks, a more extensive comparison of these models is still needed within the Turkish NLP research community. In this paper, we focus on generating and evaluating static word embeddings for Turkish by utilizing various approaches which are Word2Vec Skip-gram, Word2Vec CBOW, FastText, GloVe, BERT, and ELMo. To create the Word2Vec, FastText, and GloVe embeddings, we used a dataset composed of two Turkish corpora: \textit{BounWebCorpus} \cite{20.500.12913/16} and \textit{HuaweiCorpus} \cite{yildiz2016morphology}. For ELMo and BERT, we utilized publicly available pre-trained models. We employed two different strategies for converting contextual embeddings into static embeddings. The first approach collects the embeddings of a word in different contexts and applies a pooling operation \cite{bommasani-etal-2020-interpreting}. The second approach, on the other hand, integrates the contextual information into a static model \cite{gupta-jaggi-2021-obtaining}. We evaluate the quality of these embeddings through analogy and similarity tasks, as well as their performance on various NLP tasks which are binary sentiment analysis, named entity recognition, and part-of-speech tagging.

While contextual word embeddings have demonstrated remarkable success across various NLP applications, it is crucial to acknowledge the continued significance of static word embeddings. We firmly assert the importance of analyzing static word embeddings due to the following reasons: (1) As pointed out by \cite{gupta-jaggi-2021-obtaining}, the utilization of static word embeddings incurs a computational cost that is typically tens of millions of times lower when compared to the usage of contextual embedding models. We strongly believe that, as put by \cite{strubell-etal-2019-energy} in their work for energy and policy considerations for deep learning, researchers should prioritize computationally efficient hardware and algorithms. Analysis of static word embeddings for Turkish, especially generating static embeddings from contextual models, provides researchers and practitioners with an alternative option that requires less energy. (2) As also stated by \cite{gupta-jaggi-2021-obtaining}, many NLP applications require usage of static word embeddings by nature as evidenced by research in interpretability, bias detection and removal, word vector space analysis, and other non-contextual metrics \cite{kaneko-bollegala-2019-gender, DBLP:journals/corr/abs-1903-03862, manzini-etal-2019-black, vulic-etal-2020-good}.

The contributions in this paper are as follows:
\vspace{-0.7em}
\begin{itemize}[leftmargin=*]
\itemsep-0.3em
    \item We generate static word embeddings from contextual models for Turkish using popular methods in the literature and analyze their performances.
    \item We extensively evaluate a large number of static word embeddings for Turkish, extracted from both contextual and non-contextual models, using intrinsic evaluation (analogy and similarity tasks). 
    \item We measure the performances of these word embeddings by using them as input features to three downstream NLP tasks: sentiment analysis, part-of-speech tagging, and named entity recognition.
    \item This is the first work in Turkish where contextual embeddings are converted to static embeddings, and contextual and non-contextual models' embeddings are evaluated and compared with each other.
    \item We make our evaluation scripts, datasets, notebooks, and selected word embeddings available in a public repository for the purpose of facilitating further research and utilization \footnote{\url{https://github.com/Turkish-Word-Embeddings/Word-Embeddings-Repository-for-Turkish}}.
\end{itemize}

The rest of the paper is organized as follows. In Section \ref{relatedworks} we provide an overview of previous studies aimed at obtaining Turkish word embeddings. In Section \ref{methodology} we explain in detail the word embedding methods used in this work and the related software environments we used. Section \ref{datasets} explains the training corpora and the evaluation datasets. This is followed by the experiments and results in Section \ref{experiments}. In Section \ref{generalization}, we discuss possible adaptations of our methodology to diverse languages and how word embeddings in Turkish compare to other languages. Finally, Section \ref{conclusion} concludes the paper.

\section{Related Works}
\label{relatedworks}
   
Various studies have been conducted to scrutinize the performance of word embeddings for the Turkish language. These investigations encompass the utilization of evaluation datasets as well as mainstream NLP tasks. 

In their study, \cite{gungor2017linguistic} investigated the effectiveness of word embeddings generated using the Word2Vec Skip-gram model in capturing information related to inflection and derivation groups in Turkish. They conducted various analogy tasks and observed varying capabilities of the embeddings. Using Word2Vec trained on \textit{HuaweiCorpus} \cite{yildiz2016morphology}, they obtained a mean reciprocal rank score of 0.547 for noun conjugation suffixes and 0.273 for verb conjugation suffixes. Based on the evaluation results, they pointed out that as Turkish words acquire more affixes, the representational capacity of the information conveyed by these affixes diminishes. This finding underscores the need for conducting further research on word representation methods that take into consideration the rich morphological structure of the Turkish language. In our research, in addition to Word2Vec, we also documented the evaluation outcomes of the models trained using FastText, which is known to offer improved embeddings for languages with complex morphological characteristics when compared to Word2Vec.

\cite{DBLP:journals/corr/abs-2002-05417} trained Word2Vec and FastText-based word embeddings, and compared their results on different intrinsic and extrinsic tasks. They also demonstrated that manipulating words' surface forms can genuinely enhance the efficacy of text representation in a morphologically rich language. 

\cite{zcukurova} conducted a comparison between Word2Vec and GloVe models, evaluating their performance on analogy tasks and the training time. Within the scope of the study, the two fundamental algorithms of the Word2Vec method, CBOW and Skip-gram, have also been analyzed. In the representation of frequently occurring words within the corpus CBOW was found to be more successful, whereas for the representation of rare words the Skip-gram algorithm was more successful. Moreover, it has been demonstrated that Word2Vec outperforms GloVe in both training speed and evaluation results.

\cite{usen2014learning} applied the Skip-gram model on a Turkish corpus and examined its performance using the question set they generated. To measure the quality of the extracted word representations, two different categories of linear tests have been conducted. The first one is focused on analogy tasks, while the second one involves identifying the semantically outlier word within a group of words. They conducted a series of experiments involving various vector sizes, and the results indicated that a dimensionality of 300 yielded the optimal performance for the given tasks.

Apart from intrinsic evaluations, several studies have been conducted utilizing Turkish word embeddings in various NLP tasks. \cite{demir2014improving} demonstrated that leveraging the continuous vector space representations of words in a semi-supervised setting is a potent approach for named entity recognition. This method can achieve state-of-the-art performance even without employing language-dependent features for morphologically rich languages such as Turkish and Czech. 

\cite{8806491} compared three word embedding models, namely Word2Vec, FastText, and ELMo, using three distinct Turkish corpora varying in size. Similar to our results, their findings indicate that FastText performs superiorly in handling noun and verb inflections, whereas Word2Vec outperforms the others in semantic analogy tasks. Moreover, the bag-of-words model surpasses most trained word embedding models in classification tasks. 

\cite{8806539} conducted a comparison of the performance of Word2Vec, Doc2Vec, and FastText embeddings using a semi-supervised learning approach for text categorization. They evaluated various classification algorithms, including naive Bayes, support vector machines, artificial neural networks, decision trees, and logistic regression. The best classification accuracy was achieved when using FastText with support vector machines. Additionally, \cite{8907027} demonstrated that the incorporation of deep learning algorithms in conjunction with the word embedding models Word2Vec, FastText, and GloVe enhances the performance of text classification systems. The experiments showed that using LSTM as a deep learning algorithm and FastText as a word embedding model significantly improves the classification accuracy when dealing with Turkish texts.

A recent study conducted by \cite{10.1007/978-3-031-09753-9_18} involved a comparison of Turkish word embeddings trained using Word2Vec, FastText, ELMo, and BERT models. These embeddings were input into an LSTM text classification model. According to the results, BERT outperformed the other methods in word representation and emerged as the most successful model for the task at hand. Another recent work, produced by \cite{10286768}, presents a comprehensive intrinsic evaluation of Turkish word embedding models Word2Vec, GloVe, and FastText with different tasks.

In addition to research efforts, various repositories have been established to provide Turkish word embeddings, incorporating implementations for Word2Vec \footnote{\url{https://github.com/akoksal/Turkish-Word2Vec}}, GloVe \footnote{\url{https://github.com/inzva/Turkish-GloVe}}, and BERT \footnote{\url{https://github.com/stefan-it/turkish-bert}} \cite{stefan_schweter_2020_3770924}. Furthermore, word embeddings for multiple languages, including Turkish, have been generated for FastText by \cite{grave2018learning} and for ELMo by \cite{che-EtAl:2018:K18-2}.

Thus far, research endeavors have primarily concentrated on comparing the performance of a limited subset of word embeddings, typically employing either a restricted set of intrinsic evaluations or focusing on extrinsic evaluations like question-answering systems or classification tasks. In contrast, our study aims to provide a comprehensive analysis of static word embeddings by encompassing both intrinsic and extrinsic evaluation tasks across all major models, including Word2Vec, GloVe, FastText, as well as static embeddings derived from BERT and ELMo. We employ diverse corpora to train our models and utilize multiple datasets for evaluation, combining those employed in distinct research studies for a more comprehensive assessment.
Despite efforts to compare static word embeddings, there exists a research gap for Turkish in analyzing the performance of static word embeddings derived from contextual models. We demonstrate that static word embeddings, obtained from contextual models using X2Static \cite{gupta-jaggi-2021-obtaining}, outperform other static word embedding models across the majority of evaluation tasks. Our findings suggest that these static word embeddings have the potential to substitute computationally demanding contextual embedding models (or current static embedding models) in various applications where computational resources are limited. Given the significant reliance of various NLP tasks on static word embeddings \cite{shoemark-etal-2019-room}, we believe that our discoveries warrant an alternative methodology for generating Turkish word embeddings.

We deem our research comprehensive by our exhaustive coverage of the latest static word embedding methodologies. Word2Vec, GloVe, and FastText models were trained on the combination of two large corpora comprising 384,961,747 tokens in total. Our intrinsic evaluation encompasses four distinct datasets, comprising a total of 33,300 query instances. Furthermore, our extrinsic evaluation extends across three diverse NLP tasks, utilizing a total of five datasets, with 341,259 sentences for training and 41,387 sentences for testing. Our NLP datasets encompass a wide array of domains, including tweets, product reviews, wiki dumps, movie comments, Turkish national newspapers, biographical texts, instructional texts, popular culture articles, and essays. Datasets and scripts used in this study can be found in our repository \footnote{\url{https://github.com/Turkish-Word-Embeddings/Word-Embeddings-Repository-for-Turkish}}. All of our experiments including both intrinsic and extrinsic evaluation results can be reproduced using the trained models, evaluation scripts, and random seeds for NLP tasks.

\section{Word Embedding Models}
\label{methodology}

In this section, we explain in detail the contextual and non-contextual word embedding models we use in this work. The non-contextual models are the two variants of Word2Vec, FastText, and GloVe, and the contextual deep learning-based models are ELMo and BERT. We also mention the libraries and frameworks we made use of for the implementations of these models. Word2Vec and FastText are prediction-based models, which model the probability of the next word given a sequence of words (as is the case with language models), whereas GloVe is a count-based model, which leverages global co-occurrence statistics in word context matrices.

One of the state-of-the-art and commonly-used word embedding models is the \textbf{Word2Vec} model, which is not a single algorithm but a collection of different approaches and optimizations used for learning word embeddings from a large amount of data. Firstly presented by \cite{mikolov2013efficient}, it was further developed to improve the quality of the vectors and the training speed \cite{mikolov2013distributed}. Word2Vec model has two variants. While the CBOW approach aims to predict a target word based on the surrounding words, the Skip-gram approach aims to predict the surrounding words based on the target word. We used negative sampling for the Skip-gram algorithm for our experiments, but provided flexibility in our scripts so that other optimization methods can be tested out easily. The objective of the Skip-gram model is to maximize the average log-likelihood of the observed words given the surrounding context words in a corpus, which is formulated as follows:
\begin{equation}
\frac{1}{T} \sum\limits_{t=1}^{T} \sum\limits_{-m \leq j \leq m, j \neq 0} \log p(w_{t+j}|w_t)
\end{equation}
where $m$ is the context size and $T$ denotes the number of words in the training corpus. The conditional probability of a surrounding word $w_{t+j}$ given target word $w_t$ is defined as:
\begin{equation} \label{eq:prob}
p\left(w_{t+j} \mid w_t\right) = \frac{\exp\left(\text{v}_{w_{t+j}}^T \text{u}_{w_t}\right)}{\sum_{i=1}^{V} \exp\left(\text{v}_{w_i}^T \text{u}_{w_t}\right)}
\end{equation} 
where $\text{v}_{w_{i}}$ and $\text{u}_{w_j}$ stand for the word embeddings for the context word $w_{i}$ and target word $w_j$, respectively, and $V$ is the vocabulary size. $\text{u} \in$ $\mathbb{R}^d$ corresponds to the input representation of the word, whereas $\text{v} \in$ $\mathbb{R}^d$ corresponds to the output representation of the word, where d is the embedding dimension. Due to the denominator, the cost of calculating the given
probability is proportional to the vocabulary size V, which is often large ($10^5 - 10^7$ terms). To speed up the algorithm, they utilize \textit{negative sampling} in their later work \cite{mikolov2013distributed}. Then the problem reduces to the minimization of the following expression:
\begin{equation}
\frac{1}{T} \sum_{t=1}^T \left[ \log \sigma(\text{v}_{w_{t+j}}^T \text{u}_{w_{t}}) + \sum_{i=1}^k \mathbb{E}_{w_i} \left[ \log \sigma(-\text{v}_{w_i}^T \text{u}_{w_{t}}) \right] \right]
\end{equation}
where $k$ stands for the number of negative samples for training and $\sigma$ is the sigmoid function.

As a variant of Word2Vec, \textbf{FastText} represents the words as character $n$-grams and introduces a score function $s$ between the words. The conditional probability in Eqn. (\ref{eq:prob}) is modified as follows:
\begin{equation}
p\left(w_{t+j} \mid w_t\right) = \frac{e^{s\left(w_{t}, w_{t+j}\right)}}{\sum_{i = 1}^{V} e^{s\left(w_{t}, w_i\right)}}
\end{equation}
The expression becomes identical to the one used for Word2Vec when $s(w_i, w_j)$ is set to $\text{v}_{w_j}^T \text{u}_{w_i}$.
\cite{bojanowski2016enriching} proceed with a dictionary of $n$-grams of size $G$. For any word $w$, $\mathcal{G}_w \subset \{g_1, ..., g_G\}$ denotes the set of $n$-grams appearing in word $w$. Each $n$-gram $g$ is associated with a vector $z_g$. Then the scoring function between the target word $w_{t}$ and the context word $w_{t+j}$ can be defined as the sum of the $n$-gram vector representations of word $w_{t}$ as follows:
\begin{equation}
s(w_t, w_{t+j}) = \sum_{g \in \mathcal{G}_{w_t}} {z_g}^T  v_{w_{t + j}}
\end{equation}

The \textbf{GloVe} (Global Vectors) model is a log-bilinear regression model that combines two major concepts: global matrix factorization (such as latent space analysis of \cite{deerwester1990indexing}) and local context window (such as the Skip-gram model of \cite{mikolov2013efficient}). The GloVe loss function aligns word vectors by comparing their dot product to the logarithm of co-occurrence values, promoting similar representations for words that commonly co-occur:

\begin{equation}
    J = \sum_{i=1}^V\sum_{j=1}^V h(X_{ij})(\text{v}_{w_{j}}^T \text{u}_{w_{i}} + b_i + c_j - log(X_{ij}))^2
\end{equation}

Here, $X$ represents the matrix of word-word co-occurrence frequencies, where $X_{ij}$ tabulates the number of times word $j$ occurs in all the contexts of word $i$. Similar to Eqn. (\ref{eq:prob}), $\text{v}_{w_{j}}$ and $\text{u}_{w_i}$ are the word embeddings for the context word $w_{j}$ and target word $w_i$, respectively, and $V$ is the vocabulary size. $b_i$ is the bias term for $w_i$ and $c_j$ is the bias term for $w_j$. The weighting function $h$ is used to give more importance to frequent co-occurrences than the rare co-occurrences which are noisy. There are possible candidates for the function $h$, the following is the one used by the original GloVe implementation. Setting $x_{\text{max}}$ to 100, they found that $\alpha = \frac{3}{4}$ proved to be the most favorable experimental value.

\begin{equation}
    h(x)= 
\begin{cases}
    (x/x_{max})^\alpha& \text{if } x < x_{max}\\
    \;\;\;\;\;\; 1& \text{otherwise}
\end{cases}
\end{equation}

\textbf{ELMo} (Embeddings from Language Models) is a language model that takes a sequence of words as input and computes the conditional probability of each word given its left and right contexts. Given a sequence of $N$ tokens, ($t_1, t_2, ..., t_N$), the task can be defined as maximizing the log-likelihood of the word predictions using both left and right contexts using a bi-directional LSTM:
\begin{equation}
\begin{aligned}
    \sum_{k=1}^{N}  \log p\left(t_k \mid t_1, \ldots, t_{k-1}; \Theta_x, \overset{\rightarrow}{\Theta}_{LSTM}, \Theta_s\right) + \\ \log p\left(t_k \mid t_{k+1}, \ldots, t_{N}; \Theta_x, \overset{\leftarrow}{\Theta}_{LSTM}, \Theta_s\right) 
    \end{aligned}
\end{equation}
Here, $\Theta_{x}$ represents the parameters used for the input token representations and $\Theta_{s}$ represents the parameters in the last dense layer used just before the softmax activation function. Lastly, $\Theta_{LSTM}$ represents the parameters used in the forward and backward LSTM layers. In ELMo, some of the weights are shared between forward and backward directions. 

In a classical forward LSTM architecture with $L$ layers, the embedding $x_k^{LM}$ of an input token $t_k$ is passed through the layers. For creating the token embeddings, different architectures can be used such as character-level CNNs\footnote{\url{https://github.com/allenai/allennlp/blob/main/allennlp/modules/elmo.py}}. At each position $k$, a context-dependent representation $\overrightarrow{h}_{k, j}^{LM}$ is generated for layer $j$, $1 \leq j \leq L$. The right arrow indicates that the flow is from the leftmost word to the rightmost word in the sequence. In the top layer, $\overrightarrow{h}_{k, L}^{LM}$ is fed to a softmax layer to predict the token $t_{k+1}$. This approach enables utilizing all the words $t_1$ to $t_k$ in the left context to predict $t_{k+1}$.

Overall, for an $L$-layer bidirectional LSTM, $2L + 1$ vectors are computed for each token position $k$: $x_k^{LM}$, $\overrightarrow{h}_{k, j}^{LM}$, $\overleftarrow{h}_{k, j}^{LM}$, $1 \leq j \leq L$. In the simplest case, we can just take the outputs of the top layer, $\overrightarrow{h}_{k, L}^{LM}$ and $\overleftarrow{h}_{k, L}^{LM}$. However, in ELMo it is preferred to compute a task-specific weighting of all layers for token $k$:
\begin{equation}
    ELMo^{task}_{k} = E(R_k; \Theta^{task}) = \gamma_{task} \sum_{j=0}^{L}  s_j^{task} \cdot h_{k, j}^{LM}
\end{equation}
${h}_{k, j}^{LM}$ is the concatenated form of the left and right hidden vectors $\overrightarrow{h}_{k, j}^{LM}$ and $\overleftarrow{h}_{k, j}^{LM}$. ${h}_{k, 0}^{LM}$ corresponds to the input token embedding $x_k^{LM}$ concatenated with itself. $s^{task}$ is the softmax normalized weights of the layers and $\gamma_{task}$ is used to rescale the whole vector which is practically useful for optimization. It is also stated by \cite{peters2018deep} that it is beneficial to add a moderate amount of dropout to ELMo and regularize it by adding $\lambda \left\lVert \mathbf{w} \right\rVert_2^2$ to the loss function.

\textbf{BERT} (Bidirectional Encoder Representations from Transformers) is a pre-trained language model based on the transformer architecture, which learns contextualized word representations by considering both the left and right contexts in a sequence simultaneously. Through a two-step pre-training process involving masked language modeling (MLM) and next sentence prediction (NSP), BERT effectively captures fine-grained semantic relationships and contextual nuances in text. By training on vast amounts of unlabeled text data, BERT has achieved impressive results on various downstream NLP tasks, including text classification, named entity recognition, and question answering. Its ability to understand language contextually has established BERT as a state-of-the-art model in NLP, enabling significant advancements especially in tasks related to in natural language understanding.

In this work, we used the Gensim \footnote{\url{https://github.com/RaRe-Technologies/gensim}} library \cite{rehurek_lrec} to generate Word2Vec and FastText word embeddings. For GloVe, we utilized the official implementation \footnote{\url{https://github.com/stanfordnlp/GloVe}} provided by the Stanford NLP Group. For ELMo, we used the ELMoForManyLangs \footnote{\url{https://github.com/HIT-SCIR/ELMoForManyLangs}} library \cite{che-EtAl:2018:K18-2}. Lastly, we used the HuggingFace's transformers framework \cite{wolf-etal-2020-transformers} to effectively incorporate pre-trained Turkish BERT models into our research. For the intrinsic evaluation of all the word embedding models used in this work, we made use of the models provided in the Gensim library.

One of the main contributions of this work is evaluating contextual and non-contextual models together and comparing their performances. This requires converting non-contextual embeddings into static embeddings that can be used in intrinsic and extrinsic evaluation tasks. We used two distinct approaches for this transformation: \textit{Aggregate} and \textit{X2Static}. In the Aggregate approach \cite{bommasani-etal-2020-interpreting}, we obtain the contextual embedding of the word $w$ in different contexts and then pool (average) the embeddings obtained from these different contexts. The X2Static method \cite{gupta-jaggi-2021-obtaining} entails using a CBOW-inspired static word-embedding approach with additional contextual information from the contextual teacher model to generate the static embeddings.

\section{Datasets}
\label{datasets}

\subsection{Turkish Corpora}
To train the non-contextual models Word2Vec, FastText, and GloVe, we used two Turkish corpora, \textit{BounWebCorpus} \cite{20.500.12913/16} and \textit{HuaweiCorpus} \cite{yildiz2016morphology}. After merging the corpora, we trained Word2Vec, FastText, and GloVe on them for 10 epochs each without any preprocessing. The context size was set to 5. For Word2Vec and FastText, the number of instances for negative sampling was set to 5. In FastText, the minimum and maximum $n$ values for character $n$-grams were set to 3 and 6, respectively, as in the original implementation. The number of iterations was set to 100 in GloVe. The resulting corpus has a size of approximately 10.5GB. Overall, it has 1,384,961,747 tokens and  1,573,013 unique tokens, excluding tokens occurring less than the minimum frequency which we set to 10.

The model we used for ELMo had been trained on the Turkish CoNLL17 corpus which consisted of 327,299 unique tokens in total and was hosted in NLPL Vectors Repository \cite{fares-EtAl:2017:NoDaLiDa}. For BERT, we used BERTurk (uncased, 32k) published by \cite{stefan_schweter_2020_3770924}. BERTurk had been pre-trained on a filtered and sentence-segmented version of the Turkish OSCAR corpus, a recent Wikipedia dump, various OPUS corpora, and a special corpus provided by Oflazer \cite{stefan_schweter_2020_3770924}. The corpus has a size of 35GB and 4,404,976,662 tokens.

For Word2Vec, FastText, and GloVe, we trained the models with an embedding dimension of 300. ELMo used 1024 and BERT used 768 as the embedding dimension.

\subsection{Intrinsic Evaluation Datasets}
\label{intrinsicdatasets}

The quality of word embedding models is assessed using intrinsic and extrinsic evaluation methods. The most commonly used intrinsic evaluation approaches are analogy and similarity tasks. In this respect, to evaluate the word embedding models used in this work, we used semantic and syntactic analogy tasks, along with semantic and syntactic similarity tasks.

The analogy task focuses on the problem of completing the missing part in the form \textit{\enquote{a is to b as c is to ...}} with analogical reasoning. Specifically, given a relation between a pair of words $a:b$ and a word $c$, we aim to identify a word $d$ that holds the same relation with $c$. To achieve this goal, we subtract the embedding vector $a$ from the vector $b$, and subsequently add the embedding vector of $c$. We then identify the word in the embedding space that is closest to the computed vector using the cosine similarity, which corresponds to the word $d$. As \cite{mikolov2013linguistic} demonstrated in their seminal work on analogical reasoning, the task of completing an analogy can be exemplified through the popular ‘king-queen’ example, which is stated as \textit{\enquote{man is to king as woman is to ...}} and whose answer is the word \textit{queen}.

In order to observe the performance of the embedding models on different types of analogies in Turkish, we considered semantic analogies and syntactic analogies separately. For semantic analogies, we used an open-source dataset \footnote{\url{https://github.com/bunyamink/word-embedding-models/tree/master/datasets/analogy}} shared by \cite{bunyamink_word_embedding_models}. An example semantic analogy is the \textit{\enquote{adam is to kral as kadın is to ...}} question (with the answer \textit{kralice}) which corresponds to the \enquote{king-queen} analogy given above (\textit{adam}, \textit{kral}, \textit{kadın}, and \textit{kralice} correspond to \textit{man}, \textit{king}, \textit{woman}, and \textit{queen}, respectively). The semantic relationships involved in the dataset are kinship, synonym, antonym, currency, city-region, and country-capital relationships. For syntactic analogies, we utilized a dataset shared by \cite{gungor2017linguistic}. An example syntactic analogy is \textit{\enquote{git is to gider as gel is to ...}} with the answer \textit{gelir} (\textit{git}, \textit{gider}, \textit{gel}, and \textit{gelir} correspond to \enquote{(to) go}, \enquote{he/she goes}, \enquote{(to) come}, and \enquote{he/she comes}, respectively). In the original dataset, the syntactic analogies were split into two categories as instances formed by verb conjugation suffixes and instances formed by noun conjugation suffixes. We further segmented each category into more specific subcategories in our work. By dividing the semantic and syntactic analogies into fine-grained groups, we delineate a more nuanced classification framework that enables us to observe the effects of different embedding models on different types of relationships. Table \ref{table:freq} displays the complete list of semantic and syntactic analogies used in this work.

\begin{table*} [t]
\centering
\small 
\setstretch{1.1} 
\captionsetup{justification=centering}
\caption{Analogy task categories}
\label{table:freq}
\begin{tabular}{cccl}

Main Category & Subcategory & \# instances  \\
\midrule

    & first-person singular present tense & 990 \\
    & second-person singular present tense  &  946\\
    & second-person plural present tense  &  946\\
    & third-person plural present tense  &  1128\\
    & imperative  &  1176 \\
    Syntactic  & present continuous tense (\textit{-yor}) & 861 \\
    (verb conjugation suffixes) & present continuous tense  (\textit{-makta}/\textit{-mekte}) & 1176 \\
    & definite past tense & 1176 \\
    & indefinite past tense & 1176 \\
    & future tense & 1176 \\
    & obligation suffix & 1128 \\
    & optative suffix & 1176 \\
    & negation suffix & 1176 \\
  \cline{3-3}
    &  & total: 14,231 \\
\hline
    & first-person singular possessive suffix & 1128 \\
    & second-person singular possessive suffix    & 1176   \\
    & third-person singular possessive suffix & 1176 \\
    & first-person plural possessive suffix & 1128 \\
    & second-person plural possessive suffix &  1081  \\
    & third-person plural possessive suffix             &   1128 \\
    Syntactic & ablative case &  1128\\
    (noun conjugation suffixes) & dative case            &  1176  \\
    & locative case & 1128  \\
    & accusative case & 1176 \\
    & genitive case & 1176  \\
    & plural suffix &  1128 \\
    & instrumental suffix & 1128  \\
    & equative suffix &  276 \\
  \cline{3-3}
    &  & total: 15,133 \\
\hline
    & kinship  &  90  \\
    & synonym  & 600 \\
    & antonym  & 600 \\
    Semantic & currency & 156  \\
    & city-region &  1344  \\
    & country-capital & 506 \\
  \cline{3-3}
    &  & total: 3,296 \\

\end{tabular}
\end{table*}

The similarity (relatedness) task focuses on the problem of determining the degree of semantic similarity or relatedness between two words. For this purpose, we used the syntactic similarity examples from the \textit{WordSimTr} dataset prepared by  \cite{ustun2018characters} and semantic similarity examples from the \textit{AnlamVer} dataset shared by \cite{ercan-yildiz-2018-anlamver}. Each dataset is formed of word pairs and similarity scores for the pairs. In our study, we normalized the similarity scores to a numeric scale ranging from 0 to 10. An instance in the \textit{WordSimTr} dataset is a pair of words that have the same conjugation suffixes. An example is the pair \textit{\enquote{kitaplardan : hikayelerden}} (\textit{kitap} (\enquote{book}) and \textit{hikaye} (\enquote{story}) are lemma forms, the suffixes \textit{lar}/\textit{ler} and \textit{dan}/\textit{den} correspond to plural and ablative cases, respectively) with normalized similarity score of 5.401. Although referred as a syntactic similarity dataset, the similarity score in fact does not only encode syntactic similarity (both words have the same morphosyntax), but also the semantic similarity between the lemma forms of the two words. In this respect, we measure how much using the same suffixes for the words map them to similar parts of the embedding space.

\begin{figure*}[t]
\centering
\includegraphics[width=.45\textwidth]{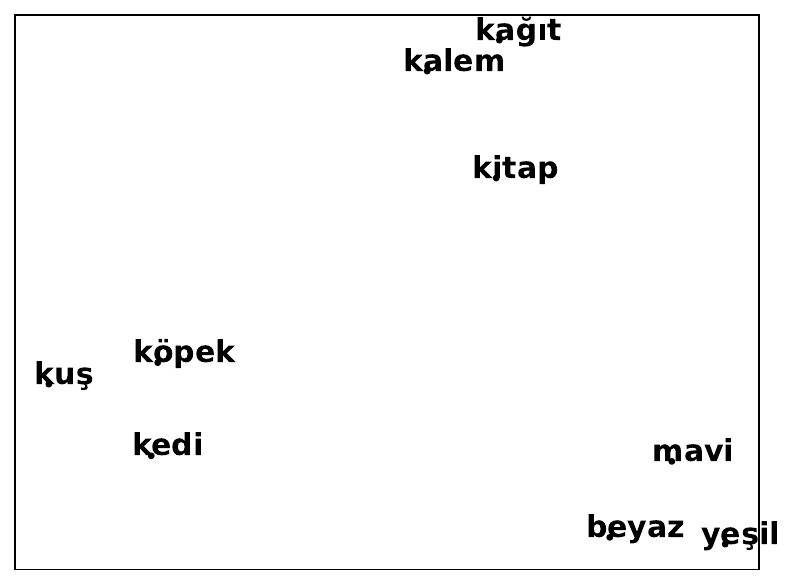}\hspace{-0.2cm}
\includegraphics[width=.45\textwidth]{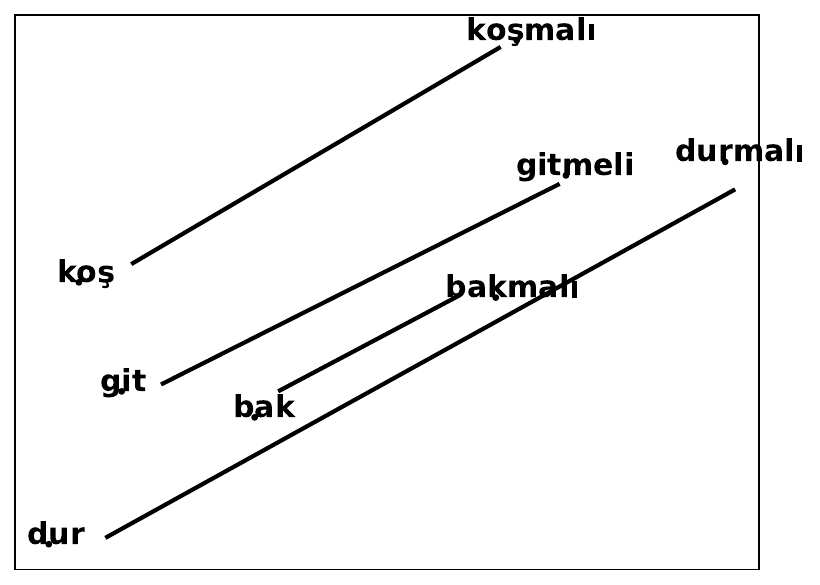}
\captionsetup{justification=centering}
\caption{Word2Vec embeddings were employed for visualizing word similarities on the left, while FastText embeddings were utilized for word analogies on the right, both employing PCA to project vectors into a two-dimensional embedding space.}
\label{fig:figure1}
\end{figure*}

Table \ref{table:intrinsic_dataset} shows the datasets, the task category covered by each dataset, and the number of instances in each dataset for the analogy and similarity tasks. Figure \ref{fig:figure1} displays a visualization\footnote{\url{https://github.com/Turkish-Word-Embeddings/Turkish-WebVectors}} of the analogy task (left) and the similarity task (right) for a few examples of Turkish words projected to two-dimensional embedding space. The left figure shows the distance between the vectors of the lemma form of a word (e.g. \textit{söyle} - \enquote{(to) say}) and its inflected form (\textit{söylemeli} - \enquote{he/she must say}). We expect a similar relationship (direction and length of the distance vector) between other lemma forms and the inflected forms with the same suffix (e.g. \textit{dur} - \enquote{(to) stop} / \textit{durmalı} - \enquote{he/she must stop}), which is mostly the case in the figure. The right figure shows word clusters indicating similarities and dissimilarities between word pairs. For instance, the embedding vectors of the words \textit{kuş} (\enquote{bird}) and \textit{köpek} (\enquote{dog}) are close to each other in the vector space produced by the model indicating a high similarity, while those of the words \textit{kuş} (\enquote{bird}) and \textit{kağıt} (\enquote{paper}) are farther away indicating a lower similarity in between.

\begin{table*}[t]
\centering
    \small 
    \captionsetup{justification=centering}
  \caption{Datasets for intrinsic evaluation tasks}
  \label{table:intrinsic_dataset}
  \begin{tabular}{cccl}
   
    Dataset&Category&\# instances\\
    \midrule
     Dataset by \cite{gungor2017linguistic} &Syntactic Analogy & 29,364\\
Dataset by \cite{bunyamink_word_embedding_models} & Semantic Analogy & 3,296\\
     \textit{WordSimTr} by \cite{ustun2018characters} & Syntactic Similarity &140\\
   \textit{AnlamVer} by \cite{ercan-yildiz-2018-anlamver} & Semantic Similarity &500\\

\end{tabular}
\end{table*}

\subsection{Extrinsic Evaluation Datasets}
\label{exteval}

As highlighted by \cite{DBLP:journals/corr/FaruquiTRD16}, the use of intrinsic evaluation in word embedding models poses certain challenges. One challenge is the subjective nature of word similarity, often mistaken for word relatedness, which can lead to incongruous assessments. Another issue is that optimizing word vectors for enhanced performance on similarity tasks can inadvertently lead to overfitting and result in non-comparable outcomes when different data splits are employed. Furthermore, the apparent efficacy of word vectors in capturing word similarity does not consistently translate to success in actual NLP problems \cite{tsvetkov-etal-2015-evaluation}. In order to address these concerns, we adopted a rigorous approach by incorporating extrinsic evaluation tasks in addition to intrinsic evaluation. In this way, we aim to obtain a more comprehensive and reliable assessment of word embedding model performance, taking into account both internal word similarities and their practical relevance in various application domains. 

We used three downstream NLP tasks to measure both the syntactic and semantic capabilities of the embedding models: sentiment analysis, part-of-speech tagging, and named entity recognition. Sentiment analysis is a semantic task while part-of-speech tagging is a syntactic task. Named entity recognition can be considered as a task in between that makes use of both syntax (e.g. capitalization) and semantics of the text. 
For sentiment analysis, we leveraged three distinct datasets. The first originates from \cite{turkmenoglu2014}, focusing on sentiment analysis within Turkish movie reviews. The second dataset encompasses product reviews and Turkish Wikipedia content, representing a diverse array of domains \cite{winvoker_turkish_sentiment_analysis_dataset}. Lastly, we utilized a dataset comprising Turkish tweets, compiled by \cite{Aydın_Güngör_2021}. We used positive and negative examples for all the datasets to perform binary sentiment analysis.  For PoS tagging, we compiled a dataset from the BOUN treebank \cite{marcsan2022enhancements} available in the Universal Dependencies website \footnote{\url{https://universaldependencies.org/}}. There are two PoS tags in the treebank where we used the Universal PoS tag. For named entity recognition, we used the dataset prepared by \cite{gungor2018improving}. The dataset involves four classes: I-ORG for organizations, I-PER for persons, I-LOC for locations, and O for non-entities. 

Table \ref{table:extrinsic_dataset} depicts the datasets used for the downstream tasks along with the number of instances in the train and test splits.

\begin{table*}[t]
\centering
    \small 
    \captionsetup{justification=centering}
 \caption{Dataset sizes (number of sentences) for extrinsic evaluation tasks}
  \label{table:extrinsic_dataset}
  \begin{tabular}{cccc}
  
    Task & Dataset & Train & Test \\
    \midrule
    Sentiment analysis & Turkish Movie Dataset \\ 
    & \cite{turkmenoglu2014} & 16,100 &  4,144 \\
    
    Sentiment analysis & Turkish Sentiment Analysis \\ 
    & Dataset \cite{winvoker_turkish_sentiment_analysis_dataset} & 286,854 & 32,873 \\
    
    Sentiment analysis & Turkish Twitter Dataset \\ 
    & Dataset \cite{Aydın_Güngör_2021} & 1,055 & 476 \\
    Named entity recognition & Turkish National Newspapers & 28,468 & 2,915 \\
    
     & with NER labels \cite{gungor2018improving} & & \\
    PoS tagging & UD BOUN Treebank \cite{marcsan2022enhancements} & 8,782 & 979 \\

  \end{tabular}
\end{table*}

\section{Experiments}
\label{experiments}

\subsection{Intrinsic Evaluation}

For the analogy task, we use the MRR (mean reciprocal rank) metric which is widely employed in evaluating systems that generate a list of potential responses to a given query. This metric, as utilized by Güngör and Yıldız in their research on Turkish word embeddings \cite{gungor2017linguistic}, provide a fine-grained evaluation of the model's performance. As explained in Section \ref{intrinsicdatasets}, for each query in the form \textit{\enquote{a is to b as c is to ...}}, we compute a vector and identify the words whose vectors are closest to this vector in the embedding space. We consider the top 10 words and rank them based on the similarities of their vectors to the computed vector. Then the reciprocal rank of the correct word (i.e. inverse of the position of the correct word in the sorted list) is taken as the score of the query. The significance of the reciprocal rank lies in its ability to reflect the relative importance of the position of the correct answer within the ranked list. The formula is given below:

\begin{equation}
MRR = \frac{1}{|Q|} \sum\limits_{i=1}^{|Q|} \frac{1}{S_i}
\end{equation}
where $Q$ represents the set of analogy queries and $S_i$ denotes the position of the correct answer in the list of closest words for the $i^{th}$ query. If the correct word does not appear among the top 10 results, we set $\frac{1}{S_i}$ to zero for that particular query.

As stated in Section \ref{methodology}, we use the Skip-gram and CBOW variants of Word2Vec, FastText, GloVe, ELMo, and BERT as the word embedding models. Table \ref{table:intrinsic1} shows the results of the analogy tasks for each model. In addition to observing the performance of the Word2Vec and FastText models separately, we also analyzed their combined performance to see the effect of combining a word embedding model with its morphology-aware variant. This is shown as \enquote{Word2Vec (SG) and FastText Average} in the table, where the embedding vector of a word is simply taken as the average of the embedding vectors in the two models.

\begin{table*}[t]
\centering
    \small
    \captionsetup{justification=centering}
  \caption{Syntactic and semantic analogy results (MRR)}
  \label{table:intrinsic1}
    \begin{tabular}{cccc}

         & \multicolumn{2}{c}{Syntactic} & Semantic\\
    \midrule
        Model & Verb Conjugation & Noun Conjugation & 
\\
        & Suffixes & Suffixes &
 \\
    \midrule

        Word2Vec Skip-gram              & \small{0.609} & \small{0.329} & \textbf{0.551} \\
        Word2Vec CBOW          & \small{0.612} & \small{0.380} & \small{0.414} \\
        FastText              & \small{0.697} & \small{0.371} & \small{0.303} \\
        Word2Vec (SG) \& FastText Average     & \small{0.658} & \small{0.368} & \small{0.470} \\
        GloVe            & \small{0.549} & \small{0.144} & \small{0.460} \\
        Aggregated ELMo            & \small{0.237} & \small{0.162} & \small{0.078} \\
        Aggregated BERT            & \small{0.038} & \small{0.036} & \small{0.034} \\
        X2Static BERT                    & \textbf{0.771} & \textbf{0.499} & \small{0.391} \\

    \end{tabular}
\end{table*}

We observe that X2Static distilled version of BERT embeddings mostly exhibits better performance than the other static embedding models. This outcome is consistent with the results obtained by \cite{gupta-jaggi-2021-obtaining} for the English language, as reported in their study. X2Static BERT outperforms the other models with a large margin in syntactic analogy, while Word2Vec continues to maintain a competitive edge to BERT in semantic analogy. By taking into account the internal structures of words, the FastText architecture proves to be useful in languages with rich morphology, which is the case for Turkish. The results show that FastText demonstrates proficiency in capturing the meaning of suffixes and thereby learning syntactic features like noun and verb inflections. However, in semantic analogy tasks, Word2Vec models outperform the FastText model \cite{bojanowski2016enriching}.

Although the Word2Vec and FastText models show relatively good performance, the GloVe model falls behind these models especially in the case of nominal suffixes. This indicates that, as GloVe is a global model, optimizing the word co-occurrences globally deteriorates the performance. The aggregated versions of the contextual models show much worse behaviour than both the non-contextual models and the X2Static conversions of the contextual models. This may be regarded as an expected result since aggregation works on a paragraph basis and assumes that all the context in a paragraph is related to the target word which may not be usually the case.

We delve into more details in Tables \ref{table:verb}, \ref{table:noun}, and \ref{table:semcat} by showing the performance of X2Static BERT, the most successful model in general, for the subcategories of syntactic and semantic analogies. The miss ratio denotes the percentage of analogy queries which failed to return a response because at least one of the words among the three words in the analogy query was missing from the repository.

\begin{table*}
\centering{
\captionsetup{justification=centering}
\caption{Syntactic analogy results for verb conjugation suffixes with X2Static BERT model}
\label{table:verb}
\begin{tabular}{ccc}

Morphological Categories & \multicolumn{1}{l}{Miss Ratio} & \multicolumn{1}{l}{MRR} \\ \midrule
first-person singular present tense & 0.116 & 0.684 \\
second-person singular present tense & 0.093 & 0.733  \\ 
second-person plural present tense & 0.075 & 0.713   \\ 
third-person plural present tense & 0.273 & 0.541 \\ 
imperative  & 0.069 & 0.779  \\ 
present continuous suffix (\textit{-yor}) & 0.015 & 0.924  \\ 
present continuous suffix (\textit{-makta}/\textit{-mekte})  & 0.044 & 0.843 \\ 
definite past tense & 0.105 & 0.750  \\ 
indefinite past tense  & 0.043 & 0.887   \\ 
future tense & 0.025 & 0.882 \\ 
obligation suffix & 0.121 & 0.710  \\ \
optative suffix & 0.079 & 0.819   \\ 
negation suffix & 0.046 & 0.751   \\ 
\hline
Overall & 0.085 & \textbf{0.771} \\
\end{tabular}}
\end{table*}

\begin{table*}[t]
\centering{
\captionsetup{justification=centering}
\caption{Syntactic analogy results for noun conjugation suffixes with X2Static BERT model}
\label{table:noun}
\begin{tabular}{ccc}

Morphological Categories & \multicolumn{1}{l}{Miss Ratio} & \multicolumn{1}{l}{MRR}  \\ \midrule
first-person singular possessive suffix  & 0.362 & 0.390 \\ 
second-person singular possessive suffix & 0.407 & 0.449  \\ 
third-person singular possessive suffix & 0.099 & 0.590  \\ 
first-person plural possessive suffix & 0.233 & 0.498  \\ 
second-person plural possessive suffix  & 0.241 & 0.527  \\ 
third-person plural possessive suffix & 0.271 & 0.327  \\ 
ablative case  & 0.171 & 0.491  \\ 
dative case  & 0.194 & 0.502 \\ 
locative case & 0.376 & 0.291  \\ 
accusative case & 0.102 & 0.646  \\ 
genitive case  & 0.093 & 0.721  \\ 
plural suffix & 0.196 & 0.599 \\ 
instrumental suffix  & 0.185 & 0.544  \\ 
equative suffix & 0.663 & 0.096  \\ 
\hline
Overall & 0.233 & \textbf{0.499} \\
\end{tabular}}
\end{table*}

\begin{table}
\centering{
\captionsetup{justification=centering}
\caption{Semantic analogy with X2Static BERT model}
\label{table:semcat}
\begin{tabular}{ccc}

Semantic Categories & \multicolumn{1}{l}{Miss Ratio} & \multicolumn{1}{l}{MRR} \\ \midrule
    kinship & 0.244 & 0.546 \\
synonyms & 0.402 & 0.390 \\ 
antonyms & 0.673 & 0.141  \\ 
currency & 0.249 & 0.439 \\ 
city-region & 0.372 & 0.437 \\ 
country-capital & 0.342 &  0.258 \\ \hline
Overall &  0.333 & \textbf{0.391} \\ 
\end{tabular}}
\end{table}

For the similarity tasks, we use the Pearson correlation coefficient and the Spearman's rank correlation coefficient for evaluation. For each query, we compute the similarity of the embedding vectors of the two words in the pair using cosine similarity and then measure how much the obtained similarity score correlates with the ground truth score given in the dataset. A high correlation indicates that the similarities and differences between the word embeddings in the model associate well with the actual similarities and differences between the word pairs in the dataset. Table \ref{table:intrinsic2} shows the results of the similarity tasks for each model. We observe a similar pattern as in the analogy tasks. The X2Static BERT embeddings mostly show the best performance. This is followed by the Word2Vec Skip-gram approach, which also outperforms BERT in semantic similarities with the Pearson correlation.

Table \ref{table:sim} displays the similarity results in a more detailed form for the X2Static BERT model. The OOV (out-of-vocabulary) ratio denotes the ratio of the words in the evaluation dataset that do not exist in the embedding repository. The OOV ratio being so high in syntactic similarities is due to the agglutinative nature (i.e. rich suffixation process) in the language. There are words in the dataset that do not appear in the training corpus despite the large size of the corpus and hence no embeddings have been learned for such words. An example is the word \textit{kedilerdenmiş} (a meaning similar to \enquote{it was one of those cats}), where the lemma \textit{kedi} (\enquote{cat}) and the suffixes \textit{ler}, \textit{den}, and \textit{miş} correspond to plural, ablative, and indefinite past tense, respectively. We attained a robust correlation between the observed similarity scores and those predicted by X2Static, yielding a statistically significant result with a $p$-value of less than $0.001$.

We conclude that the static version of BERT embeddings shows better performance than the static embeddings in both of the intrinsic evaluation tasks. Among the static embedding models, Word2Vec and FastText outperform GloVe in the context of the Turkish language. This finding aligns with the results obtained by \cite{zcukurova}.

\begin{table*}[t]
\centering
\captionsetup{justification=centering}
  \caption{Syntactic and semantic similarity results}
  \label{table:intrinsic2}
    \begin{tabular}{ccccc}
   
         & \multicolumn{2}{c}{Syntactic} & \multicolumn{2}{c}{Semantic} \\
    \midrule
        Model
        & Spearman
        & Pearson
        & Spearman
        & Pearson \\
    \midrule

        Word2Vec Skip-gram  & \small{82.53} & \small{81.68} & \small{70.77} & \textbf{74.57} \\
        Word2Vec CBOW & \small{76.89} & \small{77.62} & \small{68.76} & \small{66.15} \\
        FastText & \small{66.51} & \small{68.16} & \small{67.35} & \small{70.53} \\
        Word2Vec (SG) \& FastText Average & \small{79.12} & \small{79.19} & \small{68.15} & \small{71.98} \\
        GloVe & \small{73.45} & \small{74.13} & \small{61.59} & \small{64.33} \\
        Aggregated ELMo & \small{39.11} & \small{39.43} & \small{35.18} & \small{36.77} \\
        Aggregated BERT & \small{25.17} & \small{28.45} & \small{10.65} & \small{12.63} \\
        X2Static BERT & \textbf{84.72} & \textbf{83.03} & \textbf{75.94} & \small{70.39} \\
   
    \end{tabular}
\end{table*}

\begin{table*}[t]
\centering{
\captionsetup{justification=centering}
\caption{Syntactic and semantic similarity results with X2Static BERT model}
\label{table:sim}
\begin{tabular}{ccc}

Similarity Task                                             & \multicolumn{2}{c}{Statistics}                  \\ \midrule
 & \multicolumn{1}{c}{Pearson Correlation: 83.03}  & $p$-value: 1.708 x $10^{-21}$ \\ 
Syntactic Similarity & \multicolumn{1}{c}{Spearman Correlation: 84.72} & $p$-value: 3.966 x $10^{-23}$ \\
& \multicolumn{2}{c}{OOV Ratio: 42.86}                  \\ \hline
& \multicolumn{1}{c}{Pearson Correlation: 70.39}  & $p$-value: 1.729 x $10^{-56}$ \\
Semantic Similarity & \multicolumn{1}{c}{Spearman Correlation: 75.94} &  $p$-value: 1.721 x $10^{-70}$ \\
& \multicolumn{2}{c}{OOV Ratio: 26.20}                  \\ 
\end{tabular}}
\end{table*}

\subsection{Extrinsic Evaluation}

For extrinsic evaluation, we trained one-layer LSTM networks for the three downstream tasks. We used a hidden vector size of 16 in the LSTM models, which is a small dimension compared to the models used in the literature for similar tasks. We observed that increasing the number of hidden units in the LSTM architecture leads to closer accuracy ratios between different embedding models, which reduces the evaluation quality in distinguishing between the models. Since our goal is to observe the performance differences between the embedding models, we preferred to use a small hidden vector size. For PoS Tagging and NER tasks, we trained the models for 5 epochs with a training batch size of 32 and a validation batch size of 16. For the sentiment analysis tasks, we configured the maximum number of epochs to 15. Additionally, we implemented an early stopping criterion: if the error on the validation set begins to increase, we halt the training phase. Specifically for the third Sentiment Analysis task with the Turkish Twitter Dataset, we set the number of hidden LSTM layers to 196 and used a dropout rate of 0.4 to prevent overfitting. These modifications were implemented to address the challenge posed by the limited amount of training data available for the third task. The output of the LSTM model was fed to a layer with a sigmoid activation function for binary sentiment analysis and a softmax activation function for PoS tagging and named entity recognition. All models were trained
with an Adam optimizer (learning rate = 0.001).

A predetermined set of embedding vectors consistently yields identical results during intrinsic evaluation, given the absence of stochastic elements in the process and the pre-trained nature of the embeddings. Conversely, extrinsic evaluations inherently involve stochasticity due to the training of the LSTM architecture, even when the embeddings remain constant. To mitigate this variability, we adopt the Wilson Score Interval, a method utilized by \cite{Wang2020ACS} in a comparable examination of word embeddings, to accommodate for the inherent noise in the evaluation process. We ran each experiment three times with different random seeds (7, 24 and 30) and then applied the Wilson Interval to compute the
confidence interval for each metric at the 95\% level, assuming that the sample means are approximately normally distributed. Given the mean accuracy estimate $\hat{c}$ over different runs, its interval is calculated by \( c = \hat{c} \pm z \sqrt{\frac{\hat{c}(1-\hat{c})}{n}} \), where \( n \) is the number of observations evaluated upon (equal to the number of test sentences in our case), and \( z \) is set to 1.96 for a 95\% confidence interval.

{Table} \ref{table:extrinsic_results} shows the extrinsic evaluation results for the word embedding models. In line with the findings of the intrinsic evaluations, X2Static BERT embeddings and Word2Vec \& FastText averaged word vectors emerge as top performers across three tasks, demonstrating robust performance. This is followed closely by Word2Vec embeddings, indicating notable leadership in terms of accuracy. One difference from intrinsic evaluations is that there is not a large gap between the aggregated versions of the contextual embeddings and the other models. Even for the PoS tagging task, aggregated ELMo shows itself as one of the best-performing models. The gap tends to widen for more challenging tasks with less data available.

\begin{table*}[t]
\captionsetup{justification=centering}
  \caption{Extrinsic evaluation results}
  \centering
    \small
    \begin{adjustwidth}{1cm}{}
  \begin{tabular}{cccccc}
   
    Model & \begin{tabular}{@{}c@{}}Sentiment\\Analysis\hyperref[fn:movie]{\footnotemark[11]}\end{tabular} & \begin{tabular}{@{}c@{}}Sentiment\\ Analysis\hyperref[fn:sentiment]{\footnotemark[12]}\end{tabular} &  \begin{tabular}{@{}c@{}}Sentiment\\Analysis\hyperref[fn:twitter]{\footnotemark[13]}\end{tabular}& PoS Tagging & NER \\
    \midrule

        Word2Vec Skip-gram & 0.888 ± 0.0096 & 0.935 ± 0.0027 &0.704 ± 0.0410& 0.930 ± 0.0160 & \textbf{0.978 ± 0.0053} \\
    Word2Vec CBOW & 0.888 ± 0.0096 & 0.927 ± 0.0028 &0.702 ± 0.0411 & \textbf{0.931 ± 0.0159} & 0.975 ± 0.0059 \\
    FastText & 0.886 ± 0.0097 & 0.935 ± 0.0027 & 0.711 ± 0.0407&  0.924 ± 0.0166 & 0.977 ± 0.0053 \\
    W2V (SG) \& FT Avg. & \textbf{0.892 ± 0.0095} & 0.935 ± 0.0027 & \textbf{0.731 ± 0.0398} & 0.927 ± 0.0163 & \textbf{0.978 ± 0.0053} \\
    GloVe & 0.880 ± 0.0099 & 0.929 ± 0.0028 &0.699 ± 0.0412 &0.929 ± 0.0161 & 0.971 ± 0.0058\\
    Aggregated ELMo & 0.857 ± 0.0107& 0.909 ± 0.0031 & 0.701 ± 0.0411&0.928 ± 0.0162& 0.975 ± 0.0093 \\
    Aggregated BERT & 0.848 ± 0.0109& 0.912 ± 0.0031 & 0.669 ± 0.0423 & \textbf{0.931 ± 0.0159} & 0.977 ± 0.0054 \\
    X2Static BERT & 0.881 ± 0.0099 & \textbf{0.937+ 0.0026} &  0.687 ± 0.0417 & \textbf{0.931 ± 0.0159} & \textbf{0.978  ± 0.0053} \\
    
  \end{tabular}
  \label{table:extrinsic_results}
   \end{adjustwidth}
\end{table*}

Our study indicates that intrinsic evaluations are more effective than extrinsic evaluations in discerning between different embedding models. Intrinsic evaluations have proven to be a more dependable method for assessing model performance compared to extrinsic evaluations. This finding aligns with the observation put forth by \cite{DBLP:journals/corr/abs-1901-09785}, which suggests that the effectiveness of extrinsic evaluations largely hinges on their capacity to capture sequential information, while the significance of word meaning remains secondary in such tasks.

\section{Comparison with Other Languages}
\label{generalization}

Our methodology for benchmarking word embedding methods in the Turkish language can readily extend to other languages with similar linguistic structures. The core requirement for applying our approach is the availability of a text corpus for training the embeddings. Moreover, the existence of syntactic and semantic intrinsic tasks, such as word similarity and analogy is essential for evaluating the quality of the embeddings.

Additionally, sufficient data for conducting extrinsic tasks such as sentiment analysis, part-of-speech tagging, and named entity recognition is required for replication of our work to other languages. These extrinsic tasks serve as practical measures of the usefulness and effectiveness of the learned embeddings in downstream natural language processing applications.

We anticipate that the conclusions drawn from our study can be extrapolated to other languages sharing similarities with Turkish. Turkish is among the agglutinative languages with rich morphology, similar to Finnish, Hungarian, and Japanese. Generally, Turkish sentences adhere to the subject–object–verb sequence. Renowned for its intricate system of suffixes, Turkish poses a significant challenge for computational linguistics research due to its free word order and extensive use of affixes \cite{ASLAN201821}. Research on Finnish word embeddings conducted by \cite{venekoski-vankka-2017-finnish} indicates a similar trend to our findings in Turkish: Word2Vec Skip-Gram generally outperforms Word2Vec CBOW, and while Word2Vec excels in word intrusion tasks, FastText performs best in similarity tasks. One of the interesting findings from the same study is that, although GloVe is reported to outperform Word2Vec for the same corpus, vocabulary, window size, and training time in English \cite{pennington2014glove}, it does not perform as well compared to FastText and Word2Vec for Finnish - a trend similar to our results for Turkish. Other studies on English word embeddings show GloVe excelling in analogy tasks, while Word2Vec shines in capturing word similarity \cite{ghannay-etal-2016-word}. Our Turkish findings diverge from these English trends, underscoring the impact of language-specific nuances on embedding method effectiveness.

\section{Conclusion}

\footnotetext[11]{\label{fn:movie}Turkish Movie Dataset \cite{turkmenoglu2014}}
\footnotetext[12]{\label{fn:sentiment}Turkish Sentiment Analysis Dataset \cite{winvoker_turkish_sentiment_analysis_dataset}}
\footnotetext[13]{\label{fn:twitter}Turkish Twitter Dataset \cite{Aydın_Güngör_2021}}
\label{conclusion}

In this paper, we made a comprehensive analysis of static word embedding models for the Turkish language. The models covered are the two variants of Word2Vec, FastText, GloVe, ELMo, and BERT. To be able to compare contextual models and non-contextual models, we employed two different methods that convert contextual embeddings into static embeddings. While the first method makes use of the embeddings of the paragraphs in which the words occur, the second method integrates contextual embeddings into a CBOW-based static model during training.

We assessed the quality of the embedding models using both intrinsic and extrinsic evaluations. Intrinsic evaluation involves two main tasks: the analogy task, which aims to identify the word that exhibits a specific relationship with a given word, and the similarity task, which calculates a similarity score between pairs of words. We made a fine-grained evaluation by dividing the instances in the analogy and similarity datasets into syntactic and semantic groups and also into morphological suffix and semantic relationship categories within each group. For extrinsic evaluation, we used three downstream tasks of different syntactic and semantic natures. We built a one-layer LSTM network and fed the network with the embeddings of each embedding model to see their performance on the downstream tasks. The reason for using a simple deep learning architecture for extrinsic evaluation was to be able to directly observe the performance of the embeddings without mixing with the interior complexity of the network.

Among the contextual and non-contextual embedding models, the BERT model obtained with the X2Static approach showed the best performance in most of the evaluation tasks. This finding implies that we can use static BERT embeddings in an NLP task without the need to obtain contextual BERT embeddings for each instance of a word separately. In this way, we can bypass the substantial resource requirements entailed in fine-tuning the BERT model, while achieving improved outcomes compared to other static embedding models. The performance of BERT was mostly followed by the Word2Vec model. The FastText embeddings also yielded competitive performance, which is an expected result for Turkish due to its agglutinative nature. We observed that while the static BERT model captures syntactic features better, the Word2Vec and FastText models behave well in semantic relationships. We expect that the findings in this work may empower researchers and practitioners in making informed decisions when selecting an appropriate word embedding model for specific NLP tasks in Turkish.

Our study, benchmarking static word embedding methods using a Turkish dataset, offers valuable insights for researchers. The continued relevance of word embeddings is evident, as they remain a cornerstone in natural language processing tasks. For instance, \cite{fu-etal-2014-learning} show the effectiveness of word embeddings in semantic hierarchy construction, while \cite{WANG201812} highlight their role in biomedical NLP. Additionally, we point to \cite{doi:10.1073/pnas.1720347115} for how embeddings offer a lens into societal shifts and historical changes. Furthermore, the role of embeddings is pivotal in the work of \cite{doi:10.1080/07421222.2023.2301176}, where the authors combine human-defined sentiment word lists with word embeddings to quantify text sentiment over time, revealing evolutionary effects in financial text and their implications for business outcomes. By contributing to this field, our work empowers researchers to make informed model choices tailored to their tasks and datasets.

We integrated our models into TULAP (Turkish Language Processing Platform) \footnote{ \url{https://tulap.cmpe.boun.edu.tr/demo/trvectors}}, a language processing platform that provides resources and tools for Turkish NLP applications, to be used by language researchers and developers. Our models are conveniently accessible under the demos section of the platform. The word embeddings we generated with different models can also be found in our github repository under the release.

One line of future work may be evaluating the word embedding models in intrinsic and extrinsic settings other than the ones used in this work. Two such intrinsic methods are categorization of words into clusters based on their similarities and identifying selectional preferences between verbs and their argument nouns. The embedding models can also be tested in more complex downstream tasks such as machine translation and dialogue systems to observe their behaviour in end-to-end settings. Another future direction is making comparative analyses of non-contextual models and static versions of contextual models in other languages. We may test the embedding models in languages from different language families such as the analytical language English and the synthetic language German to compare the results with the agglutinative language Turkish. Finally, we can compare the BERT embedding model with other transformer-based models like GPT (generative pre-trained transformer) that have exhibited notable improvements in capturing word meanings and contextual information \cite{rehana2023evaluation}. However, GPT models were not included in this study due to the lack of a common method for reducing them to static word embeddings, unlike BERT, and the substantial cost associated with utilizing GPT in our research.

\section{Declaration of Generative AI Usage}

During the preparation of this work the author(s) used OpenAI ChatGPT in order to improve the language and format the LaTeX tables. After using this tool/service, the author(s) reviewed and edited the content as needed and take(s) full responsibility for the content of the publication.

\bibliographystyle{icml2023}
\bibliography{ref}

\end{document}